\documentclass{article}
\usepackage{ijcai25}

\usepackage{times}
\usepackage{soul}
\usepackage{url}
\usepackage[hidelinks]{hyperref}
\usepackage[utf8]{inputenc}
\usepackage[small]{caption}
\usepackage{graphicx}
\usepackage{amsmath}
\usepackage{amsthm}
\usepackage{amssymb}
\usepackage{booktabs}
\usepackage{multirow}  
\usepackage{color}
\usepackage[switch]{lineno}

\usepackage{appendix}
\usepackage{algorithmicx}
\usepackage{verbatim}
\usepackage{algpseudocode}
\usepackage[linesnumbered,ruled,vlined]{algorithm2e}
\usepackage{natbib}

\nolinenumbers 

\title{Boundary Prompting: Elastic Urban Region Representation \\via Graph-based Spatial Tokenization}

\author{
Haojia Zhu$^1$\and
Jiahui Jin$^1$\and
Dong Kan$^1$\and
Rouxi Shen$^1$\and
Ruize Wang$^1$\and
Xiangguo Sun$^2$\And
Jinghui Zhang$^1$
\affiliations
$^1$SouthEast University\\
$^2$The Chinese University of Hong Kong\\
\emails
\{zhuhaojia, jjin, dongkan, 213231223, 213220056\}@seu.edu.cn,
xiangguosun@cuhk.edu.hk,
Jinghui Zhang@seu.edu.cn
}
\begin{document}

\maketitle


\begin{abstract}
Urban region representation is essential for various applications such as urban planning, resource allocation, and policy development. Traditional methods rely on fixed, predefined region boundaries, which fail to capture the dynamic and complex nature of real-world urban areas. In this paper, we propose the Boundary Prompting Urban Region Representation Framework (BPURF), a novel approach that allows for elastic urban region definitions. BPURF comprises two key components: (1) A spatial token dictionary, where urban entities are treated as tokens and integrated into a unified token graph, and (2) a region token set representation model which utilize token aggregation and a multi-channel model to embed token sets corresponding to region boundaries. Additionally, we propose fast token set extraction strategy to enable online token set extraction during training and prompting. This framework enables the definition of urban regions through boundary prompting, supporting varying region boundaries and adapting to different tasks. Extensive experiments demonstrate the effectiveness of BPURF in capturing the complex characteristics of urban regions.

\end{abstract}

\section{Introduction}
Urban region representation (URR) has gained traction in recent years. It encodes urban areas into embeddings for downstream tasks, such as crime prediction~\citep{wang2017region, zhou2023heterogeneous}, check-in prediction~\citep{li2024urban, fu2019efficient}, and population prediction~\citep{li2023urban}. These embeddings serve as a critical tool for urban planning, resource allocation, and policy making by providing interpretable and actionable insights into the spatial and functional characteristics of regions.

A key aspect of urban region representation is defining what constitutes a “region” within a city. Conceptually, an urban region is an area enclosed within a defined boundary. The boundary is a polygon in the urban space encompassing various spatial entities, such as shopping centers, restaurants, and roads, along with their interactions (e.g., proximity and location relationships). To standardize comparisons and evaluations in downstream tasks, existing studies often rely on predefined, fixed boundaries of urban regions. For instance, almost all region embedding studies ~\citep{li2024urban, fu2019efficient, zhou2023heterogeneous, wu2022multi, li2023urban} train and evaluate models on datasets from New York City's Manhattan area, adhering to an official division that splits Manhattan into 180 regions. Other studies employ other official divisions of cities or use grids to partition land into uniform cells of $1 km^2$.

\begin{figure} [t]
    \centering
    \includegraphics[width=\linewidth]{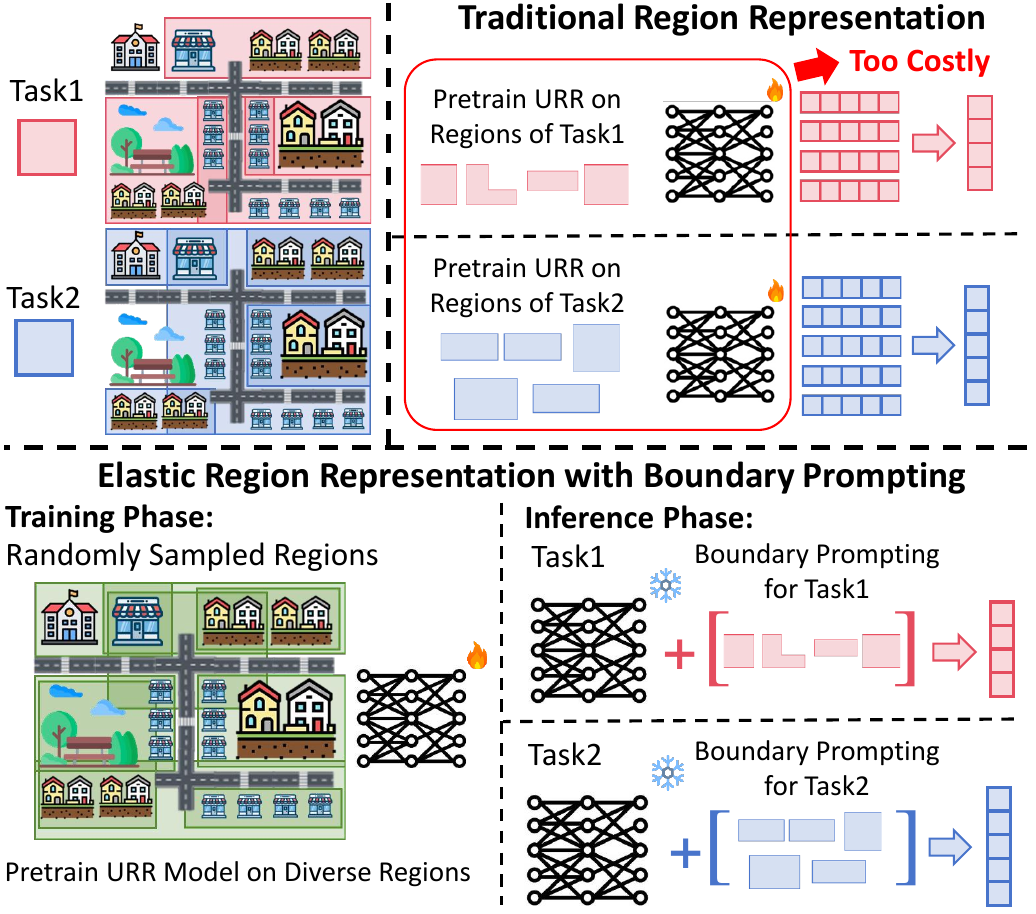}
    \vspace{-0.6cm}
    \caption{Elastic Region Representation with Boundary Prompting.}
    \label{fig:intro}
    \vspace{-17pt}
\end{figure}

However, current fixed-region approaches are far from sufficient for urban region representation because  different downstream tasks usually require varying granularities of the areas to effectively measure performance. For example, in urban livability assessments, one task might focus on services within a 10-minute walk, while another may require a broader view, such as accessibility to public transportation across a larger area. Fixed regions fail to represent such complex and dynamic nature of real-world urban areas. Besides, different urban regions may consist of overlapping spatial entities and subregions, making preset region approaches not flexible. For example, multiple commercial districts might share overlapping boundaries. Consequently, there is no single set of boundaries that can universally define how urban regions should be divided. Given this complexity, traditional approaches that rely on grid-based divisions \citep{jin2024learning, yan2024urbanclip, yuan2024unist} or static government-defined boundaries~\citep{li2024urban, zhou2023heterogeneous, li2023urban} are inadequate. As illustrated in Figure \ref{fig:intro}, when different tasks target different urban region boundaries, traditional region-based representations often require retraining the model, which is computationally expensive.


From the above discussion, we realize that a key bottleneck preventing improvements in urban region representation is the lack of support for elastic region definitions. To this end, we offer \textit{Boundary Prompting}, a novel paradigm that allows for elastic definition of urban regions. Inspired by recent advancements in prompting techniques on non-linear data \citep{sun2023all, wang2024does}, our boundary prompting treats region boundaries as a flexible input rather than a static constraint, which makes urban regions adaptive to task-specific requirements during the inference phase. 

In this paper, we propose the \textbf{B}oundary \textbf{P}rompting-based \textbf{U}rban \textbf{R}egion \textbf{R}epresentation \textbf{F}ramework (BPURF). Inspired by the token-based prompting approach in Nature Language Processing (NLP) \citep{sahoo2024systematic}, where tokens are dynamically combined based on task-specific prompts to generate context-aware representations, our framework treats urban entities as spatial tokens and urban regions as combinations of spatial tokens. During inference, the urban regions are defined through boundary prompts, which consist of multiple polygons that specify the entities to be included within each defined region. Specifically, we first introduce a general paradigm that transforms diverse urban entities into tokens and integrates them into a unified spatial token graph. Afterwards, we design a representation model that is tailored to embed any token set correspond to region boundaries. This module ensures that the structural, positional, and neighborhood information of urban regions is learned effectively, allowing the framework to adapt to diverse urban configurations. Additionally, we propose a fast token set extraction strategy to enable online token set extraction during both training and prompting. The main contributions are summarized as follows:

\begin{itemize}
\item We propose an important problem that was ignored by the research community for a long time. In this new problem, we go beyond traditional fixed-region approaches that mostly study urban representation with preset region boundaries. We push this research community forward to elastic region boundary, which will be more realistic in our real-world applications.
\item We propose a novel framework, \textit{Boundary Prompting Urban Region Representation Framework (BPURF)}, which introduces a comprehensive approach to urban region representation through a spatial token dictionary, a region token set representation module utilizing multi-channel token aggregation, and an efficient token set extraction strategy.
\item We conduct extensive experiments on multiple datasets tailored for dynamic region representation, demonstrating the superiority of BPURF in elastic urban region representation.
\end{itemize}

\section{Preliminaries}
Here, we define the concept of urban regions and introduce our goal of elastic urban region representation.

\ul{\emph{Urban entity.}} The urban entities include spatial entities, such as Point-of-Interests (POI) and roads. These entities have attributes, such as POI categories, brands, and other contextual characteristics.

\ul{\emph{Urban region.}} An urban region $r$ is fundamentally defined by its boundaries, denoted as $b_r$, which are polygons that enclose a distinct, contiguous area within a city. Each region contains the spatial entities inside the polygon.



\ul{\emph{Elastic Urban Region Representation (EURR).}} EURR problem involves learning a mapping function $\zeta: \mathcal{R} \rightarrow \mathbb{R}^{d}$ for a city. It generates a low-dimensional embedding $h_i$ for each region $r_i$. In traditional URR, the region set $\mathcal{R}$ is a finite set of predefined regions. In contrast, the region set $\mathcal{R}$ in EURR is an infinite region set that can accommodate any dynamically defined region within the city. This means that in EURR, the regions encountered during inference may not have been seen during training.

Then the mapping function can be utilized across various downstream tasks. Let the task dataset for task $\mathcal{T}$ be $\{(r_i, y_i)\}_{i=1}^{|\mathcal{R}_\mathcal{T}|}$, where $\mathcal{R}_\mathcal{T}$ is the set of regions associated with task $\mathcal{T}$ the $y_i \in \mathbb{R}$ is a numerical indicator for region $r_i$. The downstream prediction tasks aim to predict $y_i$ based on the region embedding $\zeta(r_i)$ using a simple regression model like Ridge Regression~\citep{hoerl1970ridge}.

\begin{figure} [bh]
    \centering
    \includegraphics[width=\linewidth]{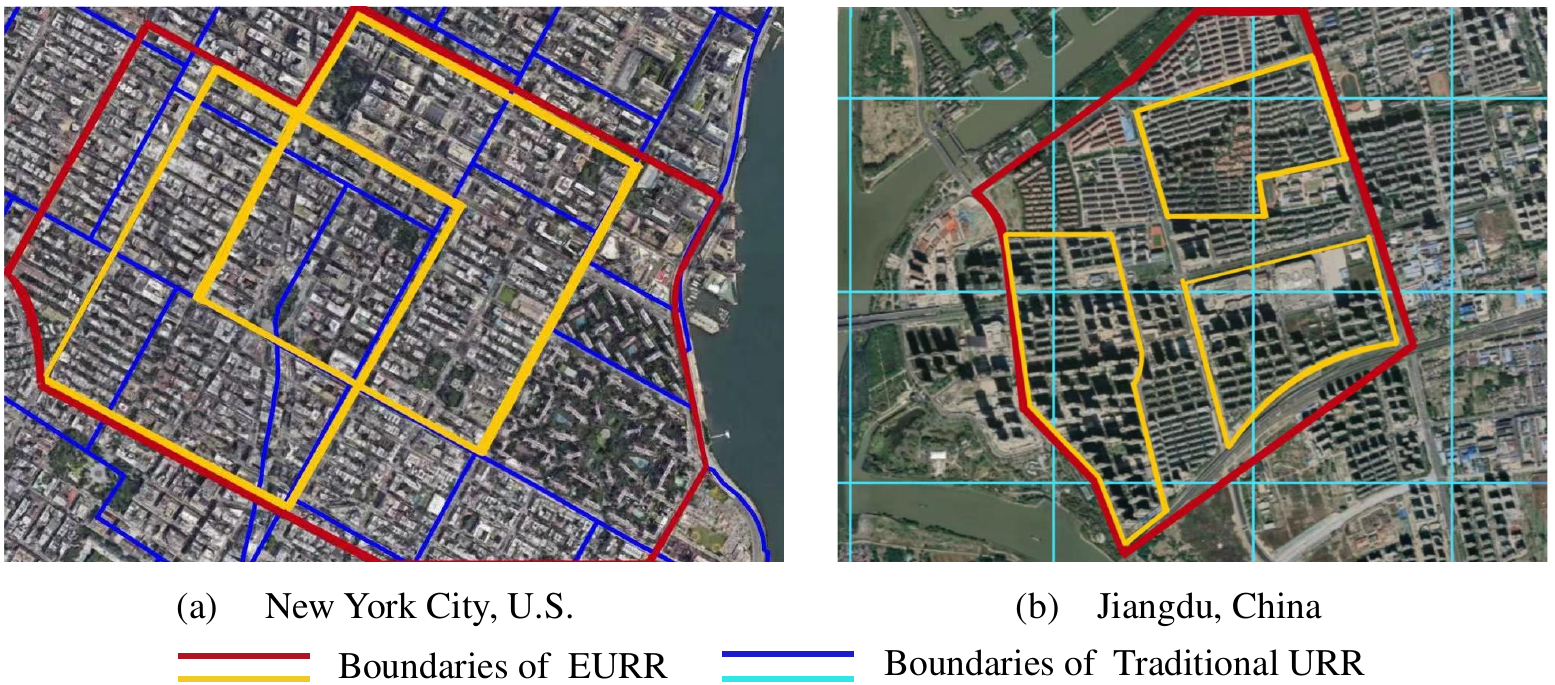}
    \vspace{-0.5cm}
    \caption{Comparison between boundaries of URR and EURR}
    \label{fig:case_study}
    \vspace{-15pt}
\end{figure}

\vspace{5pt}
{\bf {Example}.} {\it Traditional URR methods often use fixed region boundaries, such as the blue areas in Figure~\ref{fig:case_study}, representing government-defined regions in New York City or grid-based divisions in Jiangdu. However, in real world urban analysis, for tasks like urban livability assessment, the boundaries must be elastic. The yellow and red boundaries in the figure represent such elastic region definitions. The yellow boundaries correspond to smaller areas focused on specific metrics, like crime rates or population density, typically used for community-level analysis. The red boundaries represent larger regions, such as a 15-minute travel radius, which is crucial for evaluating accessibility to essential services. In such scenarios, an elastic urban region representation model is necessary.}

\setlength{\dbltextfloatsep}{5pt}
\begin{figure*}[!t]
    \centering
    \vspace{-10pt}
    \includegraphics[width=0.9\linewidth]{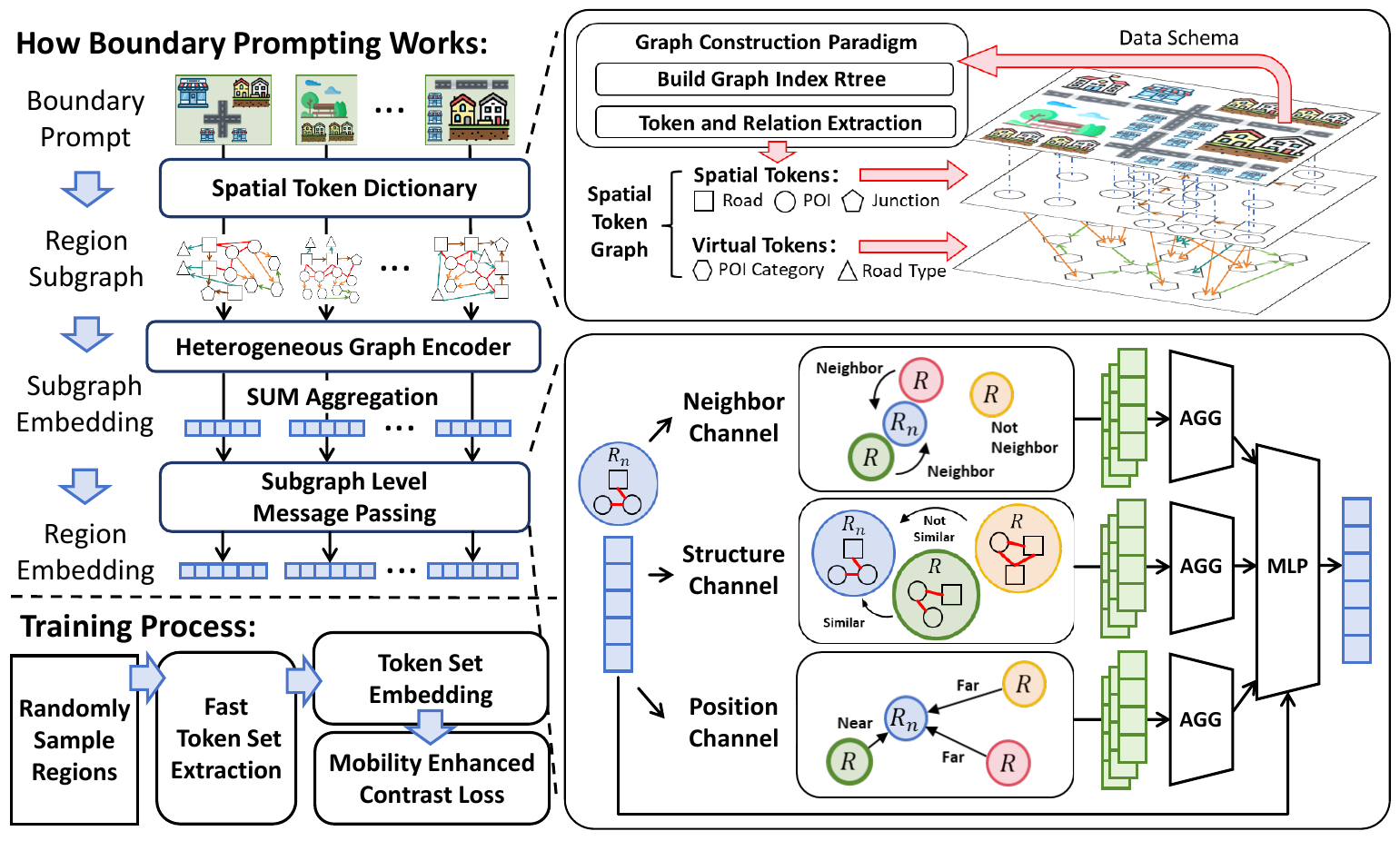}
    \vspace{-10pt}
    \caption{The core idea of BPURF is to aggregate diverse spatial entities into a unified graph, where each entity is represented as a token. A region token set representation model is then trained to map these subgraphs to region embeddings. When a boundary prompt is provided, the fast token set extraction algorithm efficiently retrieves the relevant tokens and represents them using the trained model.}
    \label{fig:Framework Overview}
\end{figure*} 

\section{Boundary Prompting Framework}

We propose BPURF, a novel framework for elastic urban region representation with boundary prompting. As shown in Figure \ref{fig:Framework Overview}, the framework first constructs a spatial token dictionary $T$ \citep{sahoo2024systematic} that represents urban entities as tokens. Then, it randomly extracts a large number of region boundaries within the city. For each boundary, the tokens related to the boundary are extracted from $T$ and combined to form a region token set. A mapping function is trained to map these token sets to region embeddings. During inference, task-specific region boundaries $\mathcal{B}=\{b_1, b_2, \dots,b_n\}$ are provided as boundary prompts, which guide the model combine the tokens to generate the final region embeddings $\mathcal{H}=\{h_1, h_2, \dots,h_n\}$. The inference process can be formulated as:
\begin{equation}
\mathcal{H} \leftarrow \mathcal{B} \oplus (T, \zeta).
\end{equation}

The first step is constructing a token dictionary $T$, which requires defining what each token represents and how it is embedded. In NLP, tokens are often derived from sequential data, and their contextual relationships can be learned from word sequences. Similarly, in our approach, tokens represent urban entities and their associated attributes. To capture the complex spatial relationships and the heterogeneous nature of entities in the urban environment, we need to embed these tokens in a way that reflects the intricate interactions between urban entities. Thus, we propose a construction paradigm that transforms urban entities and their attributes into tokens. In order to learn the initial embeddings for these tokens, we take advantage of the relationships between different urban entities to construct a token graph~(Section~\ref{section:base}).

Next, we need a model to serve as the mapping function $\zeta$ to represent urban regions. A region is associated with a set of spatial tokens. Initially, we aggregate the embeddings of these tokens to obtain a unified representation of the region. In EURR, the spatial location and internal structure of regions are not fixed, so relying solely on the aggregated token representation of one region is insufficient. To address this, we design a multi-channel message passing model that captures both the spatial and relational information between regions. This model enhances the region embeddings by enabling the regions to learn their relative positions, structural dependencies, and neighborhood relationships~(Section~\ref{section:representation}).

Finally, the efficiency of operation $\oplus$ requires optimization. This operation primarily involves retrieving the corresponding token set from the spatial token dictionary based on the boundary's polygon. During both training and inference, it is crucial to quickly identify token sets associated with region boundaries. However, since a boundary represents a polygon, finding the relevant tokens can be computationally expensive. To address this, we introduce a Fast Token Set Extraction Strategy, which accelerates the process through efficient spatial indexing and pre-computed mapping (Section~\ref{section:training}).

\section{Graph-based Spatial Tokenization} \label{section:base}
This section describes how the spatial token dictionary and graph is built. We define a paradigm to transform spatial entities and attributes into spatial tokens and virtual tokens. To obtain representations of these tokens, we leverage the relationships between different entities to construct a spatial token graph.

\subsubsection{Data Schema} \label{subsubsection:schema} 
A data schema defines a structured representation of spatial entities and relationships. Each relationship between two types of urban entities, denoted as $t_1$ and $t_2$, forms a data set $D_{t_1 \rightarrow t_2}$. Additionally, attributes are treated as virtual entities and are related to their associated entities. These attribute-entity relationships are also considered as relations within the schema. 


\subsubsection{Spatial Token Dictionary and Graph Construction} \label{subsubsection:algorithm}
We design an algorithm to transform urban entities and attributes into spatial tokens and virtual tokens, which are placed in the graph as nodes. It transform various urban entities (e.g., POIs, buildings, roads) as spatial tokens and their attributes (e.g., POI category, road types) as virtual tokens. The relationships defined in data schema are transformed into edges. More importantly, it builds an index for enabling faster retrieval of tokens corresponding to specific boundaries. The time complexity is $O\left(M \log M\right)$, where $M$ is the number of spatial tokens. More details are in supplementary files (Appendix \ref{app:graph-construction}). After constructing the token graph, we use a node embedding initialize method such as TransR to obtain the token embeddings. The spatial token dictionary consists of all the nodes in the graph and their corresponding token embeddings.



\section{Region Representation from Token Set} \label{section:representation}
Each urban region corresponds to a set of tokens from the spatial token dictionary, and the representation of a region is essentially the representation of this token set. These tokens form a region subgraph within the spatial token graph. In this section, we first describe how to aggregate the embeddings of these tokens to obtain a unified representation of the subgraph. Next, we introduce a multi-channel message passing model, which enables subgraphs to learn from each other’s spatial and relational information, capturing the interactions between different regions.

\subsection{Token Embedding and Aggregation} \label{subsection:node_agg}
We define a node encoder function $\textsf{ENC}(\cdot)$ that takes a node $v$ and its local neighborhood information as input and outputs a node embedding $h_v$. Specifically:
\begin{equation}
h_v = \text{ENC}(v, \mathcal{N}(v), G)
\end{equation}
where $\mathcal{N}(v)$ represents the neighborhood of $v$ in graph $G$. 
The subgraph embedding is then obtained by aggregating the node embeddings. In our model, the aggregation function is designed to first sum the embeddings of nodes belonging to the same type, and then concatenate the aggregated embeddings for all node types. Specifically, we group the nodes by their type and sum the embeddings within each group. Let $T_V = \{t_1, t_2, \ldots, t_k\}$ be the set of node types in the subgraph and $V_s$ be the set of nodes in the subgraph, the aggregation process for each node type $t$ is:
\begin{equation}
h_s^{(t)} = \text{SUM}\left(\{h_v \mid v \in V_s \text{ and } \phi(v) = t\}\right).
\end{equation}

To demonstrate its benefits, we provide a theoretical analysis demonstrating that our SUM aggregation function preserves regional and entity-specific information while ensuring consistent and accurate downstream predictions. We provide more detailed analysis in supplementary files (Appendix \ref{app:theory}).

\subsection{Message Passing for Token Set Embedding} \label{subsection:subgraphrelation}
We propose a message-passing mechanism tailored to region subgraphs. Then, we design multiple channels to incorporate \textbf{structural, spatial, and neighbor} information to capture different types of interactions between region subgraphs.
 


\subsubsection{Subgraph Level Message Passing}

Let set $\mathcal{X}$ represent the set of message-passing channels including structure, neighbor and position channels. For each subgraph $g_i$, the message from another subgraph $g_j$ is computed as:
\begin{equation}
\text{MSG}_{g_j \rightarrow g_i}^{(x)} \leftarrow \gamma_x(g_i, g_j) \cdot h_{g_j}
\end{equation}
where $\gamma_x(g_i, g_j)$ is the interaction weight for channel $x$, computed based on the relationship between subgraphs $g_i$ and $g_j$. The function $\gamma_x$ varies for different channels ($x$ can be structure, neighbor, or position).

The aggregated message for subgraph $g_i$ in channel $x$ is then computed using an aggregation function $\text{AGG}_M$, which collects messages from all related subgraphs $g_j \in \mathcal{R}(g_i, x)$, where $\mathcal{R}(g_i, x)$ is a function that outputs the set of subgraphs relevant to $g_i$ in channel $x$. This is expressed as:
\begin{equation}
m_{g_i}^{(x)} \leftarrow \text{AGG}_M\left(\{\text{MSG}_{g_j \rightarrow g_i}^{(x)} \mid g_j \in \mathcal{R}(g_i, x)\}\right).
\end{equation}

Finally, the embedding of subgraph $g_i$ is updated by combining the aggregated messages from all channels:
\begin{equation}
h_{g_i} \leftarrow \sigma\left(\mathbf{W} \cdot \left[h_{g_i} ; \; \{m_{g_i}^{(x)} \mid x \in \mathcal{X}\}\right]\right).
\end{equation}

Here, $\mathbf{W}$ is a learnable transformation matrix, $\sigma(\cdot)$ is a non-linear activation function, and $\{m_{g_i}^{(x)} \mid x \in \mathcal{X}\}$ denotes the concatenation of messages from all channels in $\mathcal{X}$

\subsubsection{Multi-Channel Interaction Modeling}
In this section, we design the function $\gamma$ (the interaction weight) for each of the message-passing channels. There are three channels: structure, position and neighbor channels.


For the structure channel, $\gamma_{\text{structure}}(g_i, g_j)$ is computed based on the degree distributions of nodes within the subgraphs $g_i$ and $g_j$. Specifically, for each node type in the subgraphs, we first sort the nodes by their degree. Then, we compute the Dynamic Time Warping (DTW) distance \citep{DTW} between the sorted degree sequences of $g_i$ and $g_j$. The final interaction weight $\gamma_{\text{structure}}(g_i, g_j)$ is the sum of the DTW distances for each node type:
\begin{equation}
\gamma_{\text{structure}}(g_i, g_j) = \sum_{t \in T_V} \text{DTW}(\text{Degree}_{g_i, t}, \text{Degree}_{g_j, t})
\end{equation}
where $\text{Degree}_{g_i, t}$ and $\text{Degree}_{g_j, t}$ are the sorted degree sequences for node type $t$ in subgraphs $g_i$ and $g_j$, respectively.

For the position channel, the weight $\gamma_{\text{position}}(g_i, g_j)$ is computed based on the Euclidean distance between the centers of subgraphs $g_i$ and $g_j$. The center of each subgraph is determined by the spatial coordinates of the nodes within it. To ensure the distance is on a standardized scale, we first normalize the distance between subgraph centers. The normalized distance $\tilde{d}_{g_i, g_j}$ is calculated as:
\begin{equation}
\tilde{d}_{g_i, g_j} = \frac{\|\text{Center}(g_i) - \text{Center}(g_j)\| - d_{\text{min}}}{d_{\text{max}} - d_{\text{min}}}
\end{equation}
where $d_{\text{min}}$ and $d_{\text{max}}$ represent the minimum and maximum distances among all subgraphs, respectively. The spatial weight is then computed as:
\begin{equation}
\gamma_{\text{position}}(g_i, g_j) = \exp\left(-\tilde{d}_{g_i, g_j}\right)
\end{equation}
where $\tilde{d}_{g_i, g_j}$ is the normalized Euclidean distance between the centers of subgraphs $g_i$ and $g_j$.

For the neighbor channel, the interaction weight is computed based on the proximity of the subgraphs. We assign a weight of 1 to the top $k$ closest subgraphs to $g_i$ and 0 to all others. This ensures that only the nearest subgraphs have an influence on each other in the neighbor channel.



In addition to computing the interaction weights $\gamma$, we also define the set $\mathcal{R}(g_i, x)$, which determines which subgraphs are relevant for the message-passing process in each channel. For the structure and position channels, $\mathcal{R}(g_i, x)$ includes all other subgraphs in the batch, as structural and spatial relationships are not confined to specific subgraphs but apply to all subgraphs within the input batch. For the neighbor channel, $\mathcal{R}(g_i, \text{neighbor})$ only includes the top $k$ nearest subgraphs to $g_i$, as these are the only subgraphs that are relevant in the local neighborhood context.

\subsubsection{Training Loss}
We adopt a contrastive learning approach for training. Positive samples are chosen based on spatial proximity, while negative samples are selected randomly in batch. We use InfoNCE loss \citep{gutmann2010noise} because it effectively handles this task by learning from both positive and negative sample pairs. In addition, we observe that relying solely on contrastive learning can lead to instability during training. To mitigate this, we incorporate a mobility reconstruction loss. More details can be found in supplementary files (Appendix \ref{app:loss}).


\section{Fast Token Set Extraction for Prompting} \label{section:training}
In this section, we describe how to efficiently extract the corresponding token set and region subgraph for a given boundary prompt. This process involves quickly identifying the tokens within the boundary and assembling them into a region subgraph that accurately represents the region's structure and context.




The main idea of region subgraph extraction is to first retrieve the relevant spatial tokens from the R-tree using the provided boundary. Once these spatial tokens are identified, their adjacent virtual tokens are included in the subgraph. As shown in Algorithm \ref{alg:subgraph-extraction}, it relies on two strategies: 1) \textbf{Spatial-virtual token relationship indexing} using a hashmap-based index that tracks the relationships between spatial and virtual tokens, enabling quick access to the virtual tokens connected to any given spatial token. 2) \textbf{Bitmap sampling for virtual tokens}, where a bitmap representation is used to track whether a virtual token is part of the current subgraph and avoid duplication. We use bitmap instead of a hashset because it offers better space efficiency and constant-time operations for membership tracking. Since virtual token IDs are contiguous and sequential, the bitmap allows for compact storage and fast bitwise operations to update and check inclusion status.





The overall time complexity for subgraph extraction is $O(M_q \cdot D_v)$, where $M_q$ is the number of spatial tokens within the queried boundary and $D_v$ is the average number of virtual tokens connected with per spatial token. A more detail algorithm and analysis is in the supplementary files (Appendix \ref{app:complexity}).

\section{Experiments}

We comprehensively evaluate our model BPURF, demonstrating its effectiveness across various tasks and scenarios.

\subsection{Experimental Setup}
\noindent
\subsubsection{Datasets.} We evaluate our approach using datasets from multiple cities worldwide, including New York City (NYC), Chicago (CHI), and Shenzhen (SZ), covering various tasks such as check-ins, crime, population, nightlight, and traffic accidents. To effectively validate the generalization ability of our model with dynamic regions, we source fine-grained task data and aggregate it to align with the dynamically defined urban regions. Since region boundaries and sizes vary dynamically in EURR, leading to differences in value scales, we apply standardization to ensure comparability. For each evaluation, we randomly sample 5 batches of regions and compute the average performance. Further dataset details including statistics and source, are provided in supplementary files (Appendix \ref{app:dataset}).

\noindent
\subsubsection{Baselines.} 
We compare our model with several representative baseline models, including graph embedding methods and state-of-art urban region embedding methods. Descriptions of each baseline can be found in supplementary files (Appendix \ref{app:baselines}).


\noindent
\subsubsection{Metrics and Implementation.} We evaluate the models using standard regression metrics, including MAE, RMSE, and $R^2$. Our model includes two versions:  1) BPURF that samples subgraphs for each batch, allowing the model to train on a broader variety of regions.  2) BPURF-Mini that replace the integrated subgraph extraction and training strategy with pre-sampled batches ($N_{batch} = 8$). More implementation details for our model and parameter influences can be found in supplementary files (Appendix \ref{app:parameter}).


We construct a spatial token dictionary for each city. Further details on the data schema for each city are provided in supplementary files (Appendix \ref{app:implement}).


\begin{table*}[htbp]
  \small
  \vspace{-0.3cm}
  \resizebox{1\textwidth}{!}{%
  \begin{tabular}{c|ccc|ccc|ccc|ccc}
    \toprule
    \multirow{3}[1]{*}{Model} & \multicolumn{6}{c|}{New York City~(NYC)} & \multicolumn{6}{c}{Chicago~(CHI)} \\
    \cmidrule{2-13}
     & \multicolumn{3}{c|}{Crime Prediction} & \multicolumn{3}{c|}{Check-in Prediction} & \multicolumn{3}{c|}{Crime Prediction} & \multicolumn{3}{c}{Crash Prediction}\\
    \cmidrule{2-13}
    &
    \multicolumn{1}{c}{MAE} & \multicolumn{1}{c}{RMSE} & \multicolumn{1}{c|}{$R^2$} &
    \multicolumn{1}{c}{MAE} & \multicolumn{1}{c}{RMSE} & \multicolumn{1}{c|}{$R^2$} &
    \multicolumn{1}{c}{MAE} & \multicolumn{1}{c}{RMSE} & \multicolumn{1}{c|}{$R^2$} &
    \multicolumn{1}{c}{MAE} & \multicolumn{1}{c}{RMSE} & \multicolumn{1}{c}{$R^2$}\\
    \toprule
    TransR~\citep{lin2015learning} & 0.746 & 0.984 & 0.031 & 0.658 & 0.971 & 0.055 & 0.642 & 0.991 & 0.017 & 0.572 & 0.999 & 0.001  \\
    node2vec~\citep{grover2016node2vec} & 0.822 & 1.134 & -0.288 & 0.579 & 0.905 & 0.180 & 0.773 & 1.032 & -0.066 & 0.736 & 1.016 & -0.034 \\
    GAE~\citep{kipf2016variational} & 0.760 & 1.011 & -0.023 & 0.684 & 1.010 & -0.021 & 0.652 & 1.000 & -0.001 & 0.562 & 0.999 & 0.001 \\
    \midrule
    HREP~\citep{zhou2023heterogeneous} & 0.606 & 0.946 & 0.104 & 0.539 & 0.784 & 0.386 & 0.638 & 0.863 & 0.253 & 0.478 & 0.696 & 0.515  \\
    MGFN~\citep{wu2022multi} & 0.757 & 1.109 & -0.229 & 0.587 & 0.740 & 0.453 & 0.572 & 0.823 & 0.321 & 0.426 & 0.691 & 0.522 \\
    MVURE~\citep{zhang2021multi} & 0.567 & 0.906 & 0.180 & 0.462 & 0.658 & 0.567 & 0.654 & 0.883 & 0.218 & 0.445 & 0.713 & 0.491 \\
    ReCP~\citep{li2024urban} & 0.581 & 0.873 & 0.237 & 0.401 & 0.584 & 0.659 & 0.684 & 0.920 & 0.152 & 0.435 & 0.707 & 0.499 \\
    GURPP~\citep{jin2024urban} & 0.568 & 0.748 & 0.439 & 0.394 & 0.548 & 0.699 & 0.657 & 0.928 & 0.137 & 0.451 & 0.771 & 0.405 \\
    \midrule 
    BPURF-Mini & \underline{0.398} & \underline{0.530} & \underline{0.720} & \textbf{0.219} & \textbf{0.343} & \textbf{0.883} & \underline{0.548} & \underline{0.781} & \underline{0.389} & \underline{0.387} & \underline{0.569} & \underline{0.676} \\
    Enhance & 29.93\% & 29.14\% & 64.01\% & 44.42\% & 37.41\% & 26.32\% & 16.59\% & 15.84\% & 183.94\% & 14.19\% & 26.19\% & 66.91\% \\
    BPURF & \textbf{0.376} & \textbf{0.504} & \textbf{0.745} & \underline{0.259} & \underline{0.375} & \underline{0.859} & \textbf{0.457} & \textbf{0.645} & \textbf{0.583} & \textbf{0.381} & \textbf{0.531} & \textbf{0.717} \\
    Enhance & 33.80\% & 32.62\% & 69.70\% & 34.26\% & 31.57\% & 22.89\% & 30.44\% & 30.49\% & 325.54\% & 15.52\% & 31.12\% & 77.03\% \\
    \bottomrule
\end{tabular}
}
\caption{Main performance comparison of different models on downstream tasks across NYC and CHI datasets. The best results for each metric are highlighted in bold, the second-best are underlined. The last and the third-to-last rows show the percentage improvement of our models over the third-best model.}
  \label{tab:main_performance_result}
\end{table*}

\subsection{Overall Performance}
We compare our model with several baseline models on multiple downstream tasks. To ensure fairness in the comparison, we evaluate the performance not only on dynamic regions but also on traditional fixed-region-based approaches. 1) For dynamic regions experiment, since the baseline models are not designed to work with dynamic urban regions, they can not represent new regions without extra training process. To address this, we first represent the fixed regions in the training set and then, for any new region in the validation set, we identify the top-k nearest regions based on geographical distance. Then we use a weighted sum of their representations to predict the values for the new region. 2) For the fixed-region evaluation, we follow the conventional method of training on predefined static regions and evaluating the model on the same fixed regions. This provides a comparison to the baseline methods in the context of traditional urban region representation.

The results on dynamic regions, as shown in Table \ref{tab:main_performance_result} and Table \ref{tab:shenzhen} (Appendix \ref{app:shenzhen} in supplementary files), indicate that our model outperforms baseline models across all metrics (MAE, RMSE, and $R^2$) in all datasets. From these results, we observe that existing methods, whether basic graph-based models or more specialized approaches in the field, perform poorly because they are not designed to handle unseen regions in the training process. In contrast, our method achieves significantly better performance. Compared to the best-performing baseline models, our method BPURF shows an average improvements of 30.53\% in MAE, 29.23\% in RMSE, and 35.54\% in $R^2$, demonstrating the effectiveness of our dynamic region representation.

Additionally, BPURF outperforms BPURF-Mini slightly, as it is trained on dynamically sampled regions during every batch. This enables the model to see a broader range of urban spatial features across the city, which further enhances its performance. This result supports the idea that our model benefits from exposure to a wider variety of spatial characteristics, and as the number of samples increases, its performance continues to improve. This aligns with the scaling law observe in data-driven tasks, confirming the robustness and scalability of our model. The comparison also highlights an important observation about static region-based methods, which typically train and evaluate on fixed regions. These methods risk overfitting to the specific features of those regions, limiting their ability to generalize to new, unseen areas.

While our model is designed to handle dynamic regions, it also performs exceptionally well on fixed-region tasks. As seen in Table \ref{tab:main_performance_result_static} (Appendix \ref{app:static} in supplementary files), even when evaluated on fixed regions not encountered during training, our model achieves performance comparable to or better than existing state-of-the-art (SOTA) methods.

\subsection{Scalability and Efficiency}
\begin{figure} [h]
    \centering
    \includegraphics[width=\linewidth]{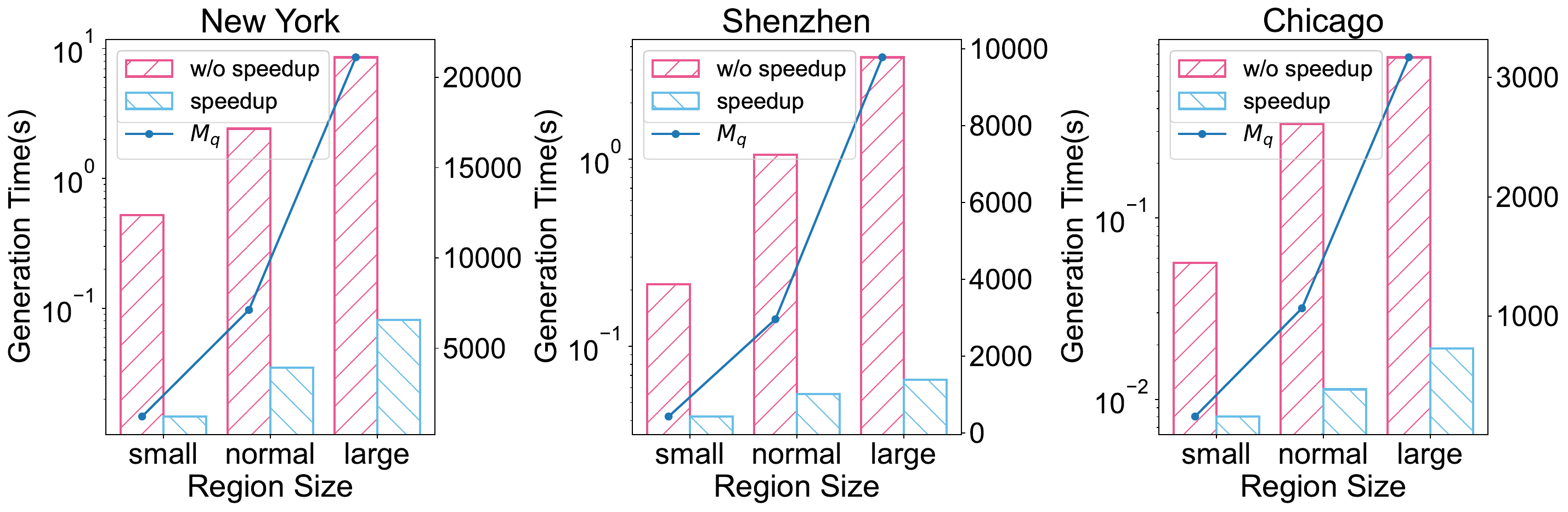}
    \vspace{-0.6cm}
    \caption{Average generation time of each region subgraph.}
    \label{fig:speedup}
    \vspace{-10pt}
\end{figure}

Our fast token set extraction strategy enables region subgraph extraction at each epoch during training. To evaluate the effectiveness of our approach, we test the average generation time for each subgraph. As shown in Figure \ref{fig:speedup}, the subgraph extraction process is accelerated by a factor of 1 to 1.5 orders of magnitude across different city scales.

Furthermore, we analyze the number of spatial points within the extracted subgraphs $M_q$ and observe that our algorithm maintains stable performance across various region subgraph sizes, regardless of the scale of the urban region. As previously mentioned, both the model complexity and the extraction complexity are closely related to $M_q$. Our experiments show that the value of $M_q$ remains within an acceptable range, ensuring that its impact on computational complexity is minimal.

\subsection{Ablation Study}
Several key components we designed in BPURF need to be validated to ensure their effectiveness: first, whether the node encoding and aggregation method (Section \ref{subsection:node_agg}) works as expected, and second, whether the components within the multi-channel interaction model (Section \ref{subsection:subgraphrelation}) are effective.

\subsubsection{Evaluation of Node Encoding and Aggregation Method}
To embed spatial tokens, we use a heterogeneous graph encoder. As part of the evaluation, we explore alternative node encoders to assess their impact on the final performance. Specifically, we test two different heterogeneous graph embedding models: HGT \citep{hu2020heterogeneous} and RGCN \citep{schlichtkrull2018modeling}. From the results in Table \ref{tab:aggregation_results}, we find that even with different node encoders, the core framework remains effective, proving the flexibility of the approach.

\begin{table}[ht]
  \small
  
  \resizebox{0.5\textwidth}{!}{%
  \begin{tabular}{c|c|ccc|ccc}
    \toprule
    \multirow{2}[1]{*}{Model} & \multirow{2}[1]{*}{AGG} & \multicolumn{3}{c|}{Crime Prediction~(NYC)} & \multicolumn{3}{c}{Check-in Prediction~(NYC)}\\
    \cmidrule{3-8}
    & &
    \multicolumn{1}{c}{MAE} & \multicolumn{1}{c}{RMSE} & \multicolumn{1}{c|}{$R^2$} &
    \multicolumn{1}{c}{MAE} & \multicolumn{1}{c}{RMSE} & \multicolumn{1}{c}{$R^2$} \\
    \toprule
    \multirow{3}[1]{*}{HGT} & SUM & 0.398 & \textbf{0.530} & \textbf{0.720} & \textbf{0.219} & \textbf{0.343} & \textbf{0.883} \\
    & MEAN & 0.533 & 0.706 & 0.502 & 0.343 & 0.530 & 0.719 \\
    & CONCAT & 0.382 & 0.559 & 0.688 & 0.238 & 0.389 & 0.848 \\  
    \toprule
    \multirow{3}[1]{*}{RGCN} & SUM & 0.409 & 0.558 & 0.689 & 0.241 & 0.364 & 0.867 \\  
    & MEAN & 0.522 & 0.703 & 0.505 & 0.371 & 0.563 & 0.683 \\  
    & CONCAT & \textbf{0.380} & 0.568 & 0.678 & 0.227 & 0.401 & 0.839 \\  
    \bottomrule
  \end{tabular}
  }
  \caption{Performance of BPURF with different node encoding and aggregation models.}
  \label{tab:aggregation_results}
  \vspace{-0.1cm}
\end{table}

We also need to validate whether the token aggregation method is effective. We replace the SUM aggregation function with two alternative methods: CONCAT and MEAN aggregation. Function CONCAT concatenates the embeddings of all nodes of each type, preserving the individual information of each node. Function MEAN computes the mean of the node embeddings of each node type. The results are shown in Table \ref{tab:aggregation_results}, where we observe that both CONCAT and MEAN result in worse performance compared to the SUM aggregation. This confirms the effectiveness of our SUM aggregation method, which satisfies the distributive property and ensures consistent downstream predictions.

\subsubsection{Evaluation of Token Set Embedding}
We evaluate the following variants of the BPURF model: “BPURF/F” removes the mobility enhancement in the training process; “BPURF/S” removes the whole subgraph level message passing; “BPURF/SP”, “BPURF/SS”, and “BPURF/SN” remove position, structure, and neighbor channels respectively.

\begin{figure} [h]
    \centering
    \vspace{-0.3cm}
    \includegraphics[width=\linewidth]{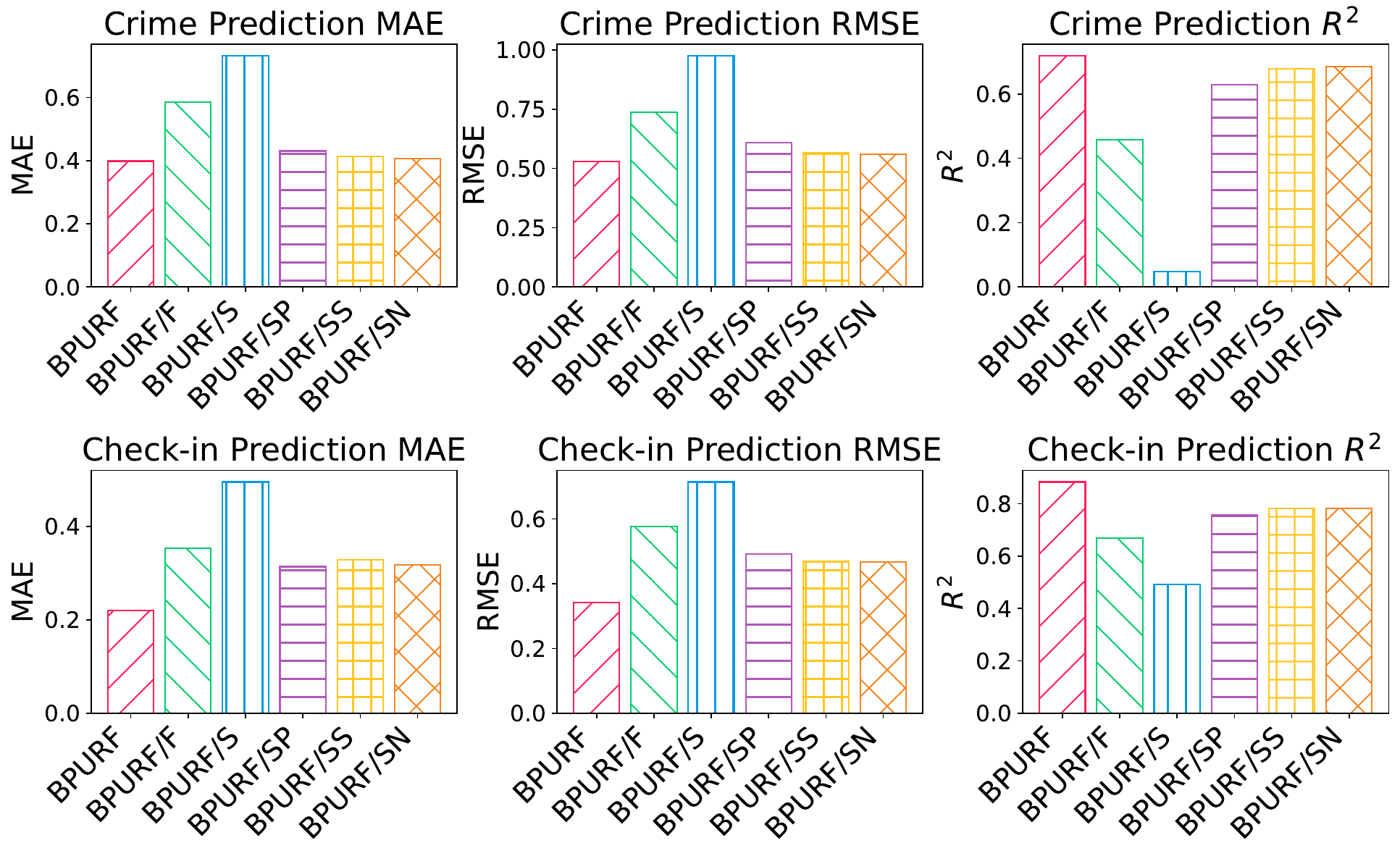}
    \vspace{-0.7cm}
    \caption{Results of ablation studies in NYC.}
    \label{fig:Ablation}
    \vspace{-13pt}
\end{figure}

From Figure \ref{fig:Ablation} and Figure \ref{fig:Ablation_chi} (Appendix \ref{app:ablation} in supplementary files), we observe that removing either the flow mechanism or the subgraph module causes a significant performance drop, with the subgraph module contributing to a larger decrease in accuracy. This demonstrates the effectiveness of our model in learning the spatial and structural relationships between subgraphs, which is crucial for dynamic region representation. In particular, removing each channel within the subgraph module results in varying degrees of performance degradation, ranging from 10\% to 30\%.


\section{Related Work}
Urban region representation has garnered significant attention in recent years, with numerous studies focusing on embedding urban regions for a variety of downstream applications.

\textbf{Urban Representation Models.} We categorize existing works into two primary categories: graph-based methods and non-graph-based methods. For non-graph methods \citep{yang2022classifying,zhou2023predicting,li2023urban,zhang2022region,li2024urban,yao2018representing,jenkins2019unsupervised}, they aggregate various data of urban entities into statistical vectors that represent different aspects of the region. For example, some works do statistics on POIs \citep{zhou2023heterogeneous}, building clusters \citep{li2023urban} and human mobility data \cite{li2024urban} to get statistical vectors of these urban entities. For graph-based methods \citep{xu2023urban,zhang2024towards,bai2019spatio,han2020stgcn,zhang2021multi,wu2022multi,liu2023urbankg,ning2024uukg}, they construct graphs with regions as nodes and create edges between nodes through various statistical metrics. For example, some works do statistics on human mobility \citep{wang2017region, yao2018representing}, poi category distribution \citep{zhou2023heterogeneous} and other region attributes and turn the statistics into edges of different graphs. Both categories are constrained by the need for predefined region boundaries and use statistical techniques for encoding urban entities. The minimum granularity in these methods is region. Thus, existing approaches cannot dynamically support arbitrary or task-specific urban region definitions.

\textbf{Prompting Methods.} There have been several studies incorporating prompting methods into urban region representation model. Some recent works \citep{zhou2023heterogeneous, jin2024urban} utilize prompts to guide models toward task-specific features, adapting embeddings to better suit particular applications. In contrast, our boundary prompting introduces a novel approach that redefines entire urban regions dynamically, offering a fresh perspective that does not conflict with these prior methods. Additionally, some studies in urban field \citep{yuan2024unist, li2024urbangpt} have applied prompting in temporal prediction tasks, which differ from our focus on region embedding. Notably, some recent work \citep{yan2024urbanclip} leverage the capabilities of large models in vision and NLP for prompting, applying them to urban region tasks. In contrast, our approach focuses primarily on spatial data, highlighting a different methodological perspective.

\vspace{-5pt}
\section{Conclusion}

In this work, we address the critical limitation of fixed-region boundaries in traditional urban region representation methods by proposing the Boundary Prompting Urban Region Representation Framework (BPURF). BPURF introduces an innovative approach to elastic urban region representation, leveraging spatial tokens and boundary prompts to support elastic region definition. By constructing a spatial token dictionary, designing a region token set representation module, and introducing an efficient token set extraction strategy, BPURF enables elastic modeling of urban regions. Extensive experiments on multiple datasets demonstrate the effectiveness and efficiency of our framework. The flexibility and scalability of BPURF enable it to handle regions of varying boundaries, making it highly suitable for a range of urban applications. Future work could explore integrating BPURF with large models to enhance its capabilities, incorporating temporal data to capture dynamic urban patterns, and enabling cross-city transfer to generalize region representations across diverse urban environments

\bibliography{ijcai25}
\bibliographystyle{named}
\clearpage
\appendix
\newpage
\section{Appendix}
\subsection{Appendix of data schema} \label{app:schema}
The full data schema $\mathcal{R}$ encompasses all relationships between different combinations of urban entity types. Each relationship $R_{t_1 \rightarrow t_2}$ is represented by a set of entity pairs $\left(e_1, e_2\right)$, where $e_1$ and $e_2$ are entities of types $t_1$ and $t_2$, respectively. We define $D_{t_1 \rightarrow t_2}$ as the data set for relationship $R_{t_1 \rightarrow t_2}$, which can be written as:
\begin{equation}
D_{t_1 \rightarrow t_2} = \left\{\left(e_1, e_2\right)|\tau\left(e_1\right)=t_1, \tau\left(e_2\right)=t_2 \right\},
\end{equation}
where $\tau$ is a function that assigns a type to each entity. The complete data set $\mathcal{D}$ is the collection of all such datasets for different combinations of entity types, denoted as:
\begin{equation}
\mathcal{D} = \left\{ D_{t_1 \rightarrow t_2} \mid t_1, t_2 \in T_V \right\}.
\end{equation}

In addition, attributes of entities are treated as virtual entities, and their relationships to the main entity are regarded as part of the schema, similar to other entity-entity relationships.

\subsection{Appendix of entity graph construction} \label{app:graph-construction}
In Algorithm \ref{alg:graph-construction}, the process begins with initializing sets for nodes $V$, edges $E$, token types $T_V$, relationship types $T_E$, a map $ID\_map$ for storing token-to-ID mappings and a graph index R-tree $GIR$. For each data tuple $D$ in $\mathcal{D}$, the algorithm performs the following operations:
1) Token Type Detection (line \ref{line:relation-loop}-\ref{line:end-typedetection}). The algorithm iterates over the pairs of related entities $(e_1, e_2)$, determining their types $t_1$ and $t_2$ and adding them to the set $T_V$ of token types. 2) Unique ID Assignment (line \ref{line:add-e1-id}-\ref{line:end-add-id}). If an token has not been assigned a unique ID, the algorithm generates one based on the current size of the corresponding set in $V$ and adds the token to the appropriate type’s set. 3) R-tree Insertion (line \ref{line:spatial-e1}-\ref{line:end-spatail-insert}): If an token is a spatial token, its coordinates and id are inserted into the R-tree $GIR$.   4) Edge Creation and Relationship Type Assignment (line \ref{line:edgecreation}-\ref{line:end-edgecreation}). Once both tokens are processed, the edge between them is created and added to the edge set $E$. After processing all pairs, the relationship type between the tokens is recorded in $T_E$.

\begin{algorithm}[h]
\caption{Token Graph Construction Algorithm} \label{alg:graph-construction}
\SetKwInOut{Input}{Input}\SetKwInOut{Output}{Output}
\SetKwFunction{GenerateID}{GenerateID}
\SetKwFunction{GetEntityType}{GetEntityType}
\SetKwFunction{AddToSet}{AddToSet}
\SetKwFunction{CreateEdge}{CreateEdge}
\SetKwFunction{Length}{Length}
\SetKwFunction{InsertToRTree}{InsertToRTree}

\Input{$\mathcal{D}$: Data tuples containing entities and their relationships \\
        $\mathcal{S}$: Spatial tokens with coordinates}
\Output{$G = (V, E, T_V, T_E)$: Token Graph\\
        $GIR$: Graph Index R-tree}

\BlankLine
\SetKwProg{ForEach}{for each}{ do}{end for each}

Initialize $V, E, T_V, T_E, ID\_map, SRT$ \\

\ForEach{$D \in \mathcal{D}$}{ \label{line:entity-loop}
    \ForEach{$(e_1, e_2) \in D$}{ \label{line:relation-loop}
        $t_1 \gets \tau(e_1)$ \\
        $t_2 \gets \tau(e_2)$ \\
        $T_V.\text{Add}(t_1)$ \\
        $T_V.\text{Add}(t_2)$ \\ \label{line:end-typedetection}
        
        \If{$e_1 \notin ID\_map$}{ \label{line:add-e1-id}
            $ID\_map[e_1] \gets \Length(V_{t_1}) $ \\
        }
        $id_1 \gets ID\_map[e_1]$ \\
        $V_{t_1}.\text{Add}(id_1)$\\

        \If{$e_2 \notin ID\_map$}{ \label{line:add-e2-id}
            $ID\_map[e_2] \gets \Length(V_{t_2})$ \\
        }
        $id_2 \gets ID\_map[e_2]$ \\
        $V_{t_2}.\text{Add}(id_2)$\\ \label{line:end-add-id}

        \If{$e_1$ is a spatial entity}{ \label{line:spatial-e1}
            $GIR$.Insert($\mathcal{S}[e_1], id_1$)
        }
        \If{$e_2$ is a spatial entity}{ \label{line:spatial-e2}
            $GIR$.Insert($\mathcal{S}[e_2], id_2$)
        } \label{line:end-spatail-insert}
        
        $E.\text{Add}((id_1, id_2))$ \label{line:edgecreation}
    }
    $T_E.\text{Add}(R_{t_1 \rightarrow t_2})$ \label{line:end-edgecreation}
}
\Output{$G = (V, E, T_V, T_E)$, $GIR$}
\end{algorithm}

The algorithm iterates over each data tuple $D \in \mathcal{D}$, and for each tuple, it iterates over the pairs of related entities $(e_1, e_2)$. Let $|\mathcal{D}|$ denote the number of data tuples and $|D|$ denote the number of entity pairs in each tuple. Therefore, the total number of iterations is proportional to $\sum_{D \in \mathcal{D}} |D|$, where $|D|$ is the size of each data tuple. Each operation within the inner loop, such as checking and adding tokens to hash sets (for token types and relationships) and hash maps (for token-to-ID mappings), is $O(1)$ on average. For R-tree insertions, the time complexity of inserting each spatial token is $O(\log M)$, where $M$ is the number of spatial tokens inserted into the R-tree. The total time complexity for inserting all spatial entities into the R-tree is $O(MlogM)$. Thus, the overall time complexity of the algorithm is:
\begin{equation}
O\left(\sum_{D \in \mathcal{D}} |D| + M \log M\right)
\end{equation}

\subsection{Appendix of token embedding and aggregation}
In this section we give more details of token embedding and aggregation.

\subsubsection{More details about embedding and aggregation} \label{app:agg}
We define a node encoder function $\textsf{ENC}(\cdot)$ that takes a node $v$ and its local neighborhood information as input and outputs a node embedding $h_v$. Specifically:
\begin{equation}
h_v = \text{ENC}(v, \mathcal{N}(v), G)
\end{equation}
where $\mathcal{N}(v)$ represents the neighborhood of $v$ in graph $G$. The encoder can be any heterogeneous graph encoder. We recommend Heterogeneous Graph Transformer (HGT) because it assigns unique transformation matrices to different node and edge types, enabling a tailored representation for diverse urban entities.

To obtain a subgraph embedding $h_s$, we aggregate the embeddings of all nodes within the subgraph. Let $V_s$ represent the set of nodes in the subgraph. The aggregation process for each subgraph embedding is defined as:
\begin{equation}
h_s = \text{AGG}\left(\{h_v \mid v \in V_s\}\right)
\end{equation}
where $h_v$ is the node embedding for node $v$.

In our model, the aggregation function $\textsf{AGG}(\cdot)$ is designed to first sum the embeddings of nodes belonging to the same type, and then concatenate the aggregated embeddings for all node types. Specifically, we first group the nodes by their type and sum the embeddings within each group. Let $T_V = \{t_1, t_2, \ldots, t_k\}$ be the set of unique node types in the subgraph, the aggregation process for each node type $t$ is:
\begin{equation}
h_s^{(t)} = \text{SUM}\left(\{h_v \mid v \in V_s \text{ and } \phi(v) = t\}\right).
\end{equation}

After performing the sum operation for each node type, the final region subgraph embedding $h_s$ is obtained by concatenating the aggregated embeddings of all node types:
\begin{equation}
h_s = \text{CONCAT}\left(h_s^{(t_1)}, h_s^{(t_2)}, \ldots, h_s^{(t_k)}\right).
\end{equation}

This approach ensures that the node embeddings are both aggregated by type and concatenated to capture all relevant information for the entire region subgraph.

\subsubsection{Theoretical analysis of aggregation function}  \label{app:theory}
While there are various aggregation options in previous works, such as average pooling or concatenate, our design opts for a simple yet straightforward approach—SUM aggregation. Despite its simplicity, this method offers two main benefits.

Firstly, it preserves regional and entity-specific information. The SUM aggregation method effectively preserves size-related characteristics of regions, ensuring that subgraphs of different sizes have distinct embeddings. Additionally, by aggregating nodes of the same type separately before concatenation, our approach captures the heterogeneous nature of urban entities. This enables the framework to model the varying impacts of different types of entities on the final subgraph embedding.

Secondly, our aggregation function satisfies the distributive property, which is critical for ensuring consistent downstream predictions. Let us consider a downstream task $\mathcal{T}$ that involves predicting a value $y$ for an urban region $r$ based on its node embeddings. For a region $r$, its prediction $y$ is typically computed as a linear combination of the embeddings of its constituent subregions, such as:
\begin{equation}
y = \text{AGG}(X) \beta + \epsilon
\end{equation}
where $X$ is the node embedding matrix for the region, $\beta$ is the regression coefficient vector, and $\epsilon$ is the error term. When the region $r$ is decomposed into several disjoint subregions $r_1, r_2, \dots, r_n$, with corresponding node embedding matrices $X_1, X_2, \dots, X_n$, the prediction for the entire region should be the sum of the predictions for the subregions, i.e.,
\begin{equation}
y = \sum_{i=1}^{n} \text{AGG}(X_i) \beta + \epsilon = \sum_{i=1}^{n} y_i + \epsilon
\end{equation}
where $y_i = \text{AGG}(X_i) \beta + \epsilon$ is the prediction for each subregion. In this case, the embeddings for the entire region $r$ should be consistent with the sum of the embeddings of its subregions, which can be expressed as:
\begin{equation}
\text{AGG}\left(\bigcup_{i=1}^{n} \{h_v \mid v \in V_{r_i}\}\right) = \sum_{i=1}^{n} \text{AGG}\left(\{h_v \mid v \in V_{r_i}\}\right).
\end{equation}

Our proposed aggregation method naturally satisfies this requirement, as it ensures that the embeddings of the entire region can be computed as the sum of the embeddings of the subregions. This property enables accurate and consistent predictions in downstream tasks.

\subsection{Appendix of loss function} \label{app:loss}
We adopt a contrastive learning approach for training. Positive samples are chosen based on spatial proximity, while negative samples are selected randomly. The contrastive loss is used to minimize the distance between embeddings of positive samples and maximize the distance between embeddings of negative samples. We use InfoNCE loss \citep{gutmann2010noise} because it effectively handles this task by learning from both positive and negative sample pairs in a normalized embedding space, thus ensuring meaningful representation learning. Let $\mathbf{r}_i$ be the embedding of region $i$, $\mathbf{p}_i$ be the embedding of its corresponding positive sample, and $\mathbf{n}_i$ be the embeddings of its negative samples. The InfoNCE loss is defined as:
\begin{equation}
\mathcal{L}_{\text{InfoNCE}} = - \sum_{i=1}^{N} \log \frac{\exp(\mathbf{r}_i \cdot \mathbf{p}_i / \tau)}{\exp(\mathbf{r}_i \cdot \mathbf{p}_i / \tau) + \sum_{j=1}^{K} \exp(\mathbf{r}_i \cdot \mathbf{n}_{ij} / \tau)}.
\end{equation}

In addition, we observe that relying solely on contrastive learning can lead to instability during training. To mitigate this, we incorporate a mobility reconstruction loss $\mathcal{L}_{\text{mobility}}$ and a prediction loss $\mathcal{L}_{\text{pred}}$ that encourages the model to maintain coherence between the flow-based embeddings of subgraphs.

Finally, the total loss function for training is the weighted sum of the three components:
\begin{equation}
\mathcal{L}_{\text{total}} = \mathcal{L}_{\text{InfoNCE}} + \mathcal{L}_{\text{mobility}} + \mathcal{L}_{\text{pred}}.
\end{equation}

\subsection{Appendix of token set extraction complexity} \label{app:complexity}
The main idea of region subgraph extraction is to first retrieve the relevant spatial tokens from the R-tree using the provided boundary. Once these spatial tokens are identified, their adjacent virtual tokens are included in the subgraph. As shown in Algorithm \ref{alg:subgraph-extraction}, it relies on two strategies: 1) \textbf{Spatial-virtual token relationship indexing} using a hashmap-based index (line \ref{line:index}) that tracks the relationships between spatial and virtual tokens, enabling quick access to the virtual tokens connected to any given spatial token. 2) \textbf{Bitmap sampling for virtual tokens} (line \ref{line:bitmap}), where a bitmap representation is used to track whether a virtual token is part of the current subgraph and avoid duplication. We use bitmap instead of a hashset because it offers better space efficiency and constant-time operations for membership tracking. Since virtual token IDs are contiguous and sequential, the bitmap allows for compact storage and fast bitwise operations to update and check inclusion status.

\begin{algorithm}[t]
\caption{Region Subgraph Extraction Algorithm} \label{alg:subgraph-extraction}
\SetKwInOut{Input}{Input}\SetKwInOut{Output}{Output} \SetKwFunction{ExtractSubgraph}{ExtractSubgraph}

\Input{
    $boundary$: The boundary prompt. \\
    $G$: The entire graph. \\
    $GIR$: The R-tree spatial index. \\
    $Index_{sv}$: The spatial-virtual token index.\\
    $bitmap$: A bitmap to track virtual tokens.
}

\Output{$subgraph$: The extracted region subgraph.}

\BlankLine
$subgraph \gets \emptyset$ \\
$spatial\_tokens \gets GIR.query(boundary)$ \\
$subgraph.add(spatial\_tokens)$ \\
\ForEach{$e_s \in spatial\_tokens$}{
    $virtual\_tokens \gets Index_{sv}[e_s]$ \\ \label{line:index}
    \ForEach{$e_v \in virtual\_tokens$}{
        $bitmap[e_v] \gets 1$  \label{line:bitmap}
    } 
} 
\ForEach{$e_v \in bitmap$}{
    \If{$bitmap[e_v] = 1$}{
        $subgraph.add(e_v)$
    }
}
\Return{$subgraph$}
\end{algorithm}
The overall complexity of Algorithm \ref{alg:subgraph-extraction} is low. Given the boundary, the R-tree query operation for spatial tokens has a time complexity of $O(M_q)$, where $M_q$ is the number of spatial tokens within the queried boundary. For each spatial token, we retrieve the connected virtual tokens using the spatial-virtual token index, which has a constant-time lookup for each spatial token. Additionally, the bitmap allows us to track the inclusion of virtual tokens in the subgraph with constant-time complexity per token. Thus, the time complexity for processing all virtual tokens associated with spatial tokens is $O(M_q \cdot D_v)$, where $D_v$ is the average number of virtual tokens connected with per spatial token. In total, the overall time complexity for subgraph extraction is $O(M_q \cdot D_v)$.

\subsection{Appendix of extra spatial augment process}
We present a comprehensive approach to model urban region information by extracting subgraphs that capture the relationships between spatial tokens. However, subgraphs can miss the layout structure within a region, particularly the spatial proximity and compactness of tokens. For instance, in two distinct urban areas, one might contain a dense cluster of tokens, while the other might have tokens that are spread out over a larger area. Without augmenting the spatial connections, a subgraph could fail to distinguish between these two regions, leading to inaccurate representations.

To address this limitation, we introduce a spatial augmentation technique that transforms proximity information into subgraph structural features.

The augmentation process is formalized through the following decision rule: for a given spatial token $e$, we identify its top-$k$ nearest neighbors, denoted as $N(e) = \{ e_1, e_2, \dots, e_k \}$. An edge is added between $e$ and each token $e_i$ if the distance $\text{distance}(e, e_i)$ is below a predefined maximum threshold $d_{\text{max}}$. This condition is formally expressed as:
\begin{equation}
\text{AUG}(e, e_i) = 
\begin{cases} 
\text{True} & \text{if } \text{distance}(e, e_i) \leq d_{\text{max}} \\
\text{False} & \text{otherwise}
\end{cases}
\end{equation}
This decision rule ensures that only the top-$k$ closest tokens, within a specified distance threshold, are connected, thereby preventing unnecessary connections.

For implementation, We first use R-tree to efficiently query the top-$k$ nearest neighbors for each spatial token $e$. Then, we apply the condition on the distances to determine whether to add an edge between $e$ and $e_i$ based on the threshold $d_{\text{max}}$. The complexity of this process is dominated by the R-tree query, which typically operates in $O(\log M_q)$ for a nearest-neighbor search, where $M_q$ is the total number of spatial tokens in the subgraph. For each spatial token, the query finds the top-$k$ closest neighbors in logarithmic time, and the distance checks are constant time operations, making the overall complexity for each spatial token approximately $O(k \log M_q)$.

\subsection{Appendix of dataset sources} \label{app:dataset}
We collect the datasets of region division, POI, taxi trips, road network, check-in, crime and crash for three representative cities: New York City (NYC) and Chicago (CHI) and ShenZhen (SZ). The detailed statistics and sources of the dataset is shown in Table \ref{tab:dataset}.

\begin{table} [h]
  \small
  \centering
  \resizebox{0.4\textwidth}{!}{%
  \begin{tabular}{c|c}
    \toprule
    Data & Source\\
    \midrule
    Region & Census Bureau~\href{https://www.census.gov/}{[link]}, SZ\href{https://gitcode.com/open-source-toolkit/93c46}{[link]} \\
    \midrule
    POI & OSM~\href{https://www.openstreetmap.org/}{[link]}, Safegraph~\href{https://www.safegraph.com/}{[link]}\\
    Taxi trip & NYCOD~\href{https://opendata.cityofnewyork.us}{[link]}, CHIDP~\href{https://data.cityofchicago.org/}{[link]},STL~\href{https://github.com/cbdog94/STL}{[link]}\\
    Road & OSM~\href{https://www.openstreetmap.org/}{[link]}\\
    Junction & OSM~\href{https://www.openstreetmap.org/}{[link]}\\
    \midrule
    Crime & NYCOD~\href{https://opendata.cityofnewyork.us}{[link]}, CHIDP~\href{https://data.cityofchicago.org/}{[link]}\\
    Check-in & Foursquare~\href{https://data.cityofchicago.org/}{[link]}\\
    Crash & CHIDP~\href{https://data.cityofchicago.org/}{[link]}\\
    Population & WorldPop~\href{https://hub.worldpop.org/geodata/summary?id=29818}{[link]}\\
    Nightlight & Geodoi~\href{https://geodoi.ac.cn/WebCn/Default.aspx}{[link]}\\
    \bottomrule
  \end{tabular}
  }
  \caption{Dataset Sources.}
  \vspace{-0.4cm}
  \label{tab:dataset}
  \vspace{-0.2cm}
\end{table}

\subsection{Appendix of baselines description} \label{app:baselines}
We compare our model with several representative baseline models, including graph embedding methods and urban region embedding methods. We all use \textbf{the best parameter settings} of the baselines.

\noindent I. Knowledge graph embedding methods
\begin{itemize}
    \item \textbf{TransR} \citep{lin2015learning}: A knowledge graph embedding model that learns embeddings in different spaces for entities and relationships, enabling the model to capture complex relationship patterns in heterogeneous graphs.
    \item \textbf{node2vec} \citep{grover2016node2vec}: A graph embedding method that generates node embeddings by optimizing a biased random walk, combining breadth-first search (BFS) and depth-first search (DFS) strategies to capture diverse node neighborhood structures.
    \item \textbf{GAE} \citep{kipf2016variational} A graph autoencoder that learns low-dimensional node embeddings by reconstructing the graph's adjacency matrix through an encoder-decoder architecture, incorporating variational inference to handle graph uncertainty.
\end{itemize}
\noindent III. State-of-the-art urban region embedding methods:
\begin{itemize}
    \item \textbf{HREP} \citep{zhou2023heterogeneous} designs heterogeneous region embedding module and prompt learning module for downstream tasks. 
    \item \textbf{MGFN} \citep{wu2022multi} proposes a mobility pattern joint learning module to learn regio representation through fine-grained mobility data.
    \item \textbf{MVURE} \citep{zhang2021multi} models different types of region correlations based on both human mobility and inherent region properties.
    \item \textbf{ReCP} \citep{li2024urban} utilizes multi-view contrastive prediction model to learn multiple information views from point-of-interest and human mobility data.
    \item \textbf{GURPP} \citep{jin2024urban} introduces a subgraph-centric pre-training model to capture general urban knowledge.
\end{itemize}

\subsection{Parameter settings and influences} \label{app:parameter}
\begin{figure}[h]  
    \centering  
    \begin{minipage}[t]{0.25\textwidth}  
        \centering  
        \includegraphics[width=\linewidth]{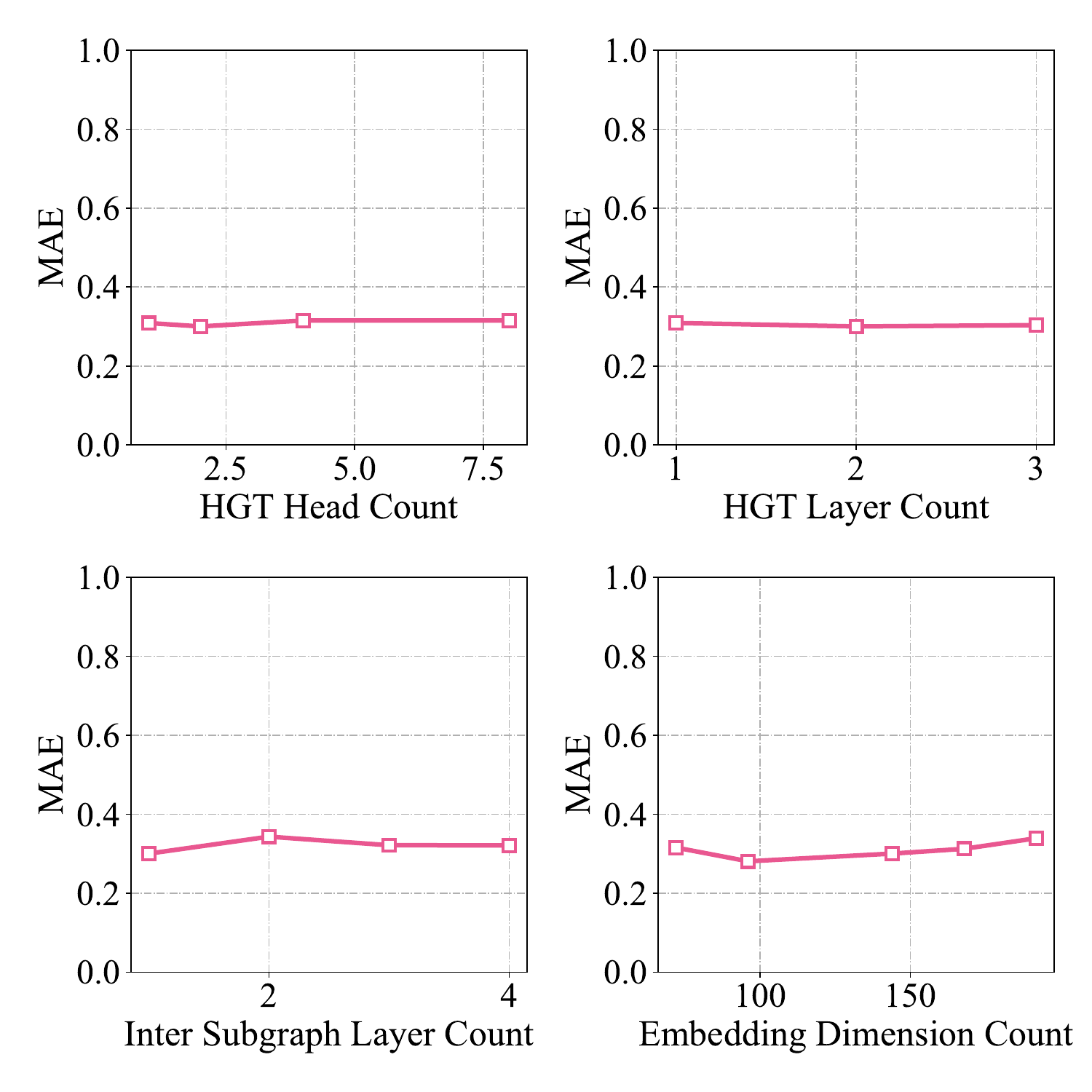} 
        \caption{New York City.}  
        \label{fig:nyc_parameter}  
    \end{minipage}
    \begin{minipage}[t]{0.25\textwidth}  
        \centering  
        \includegraphics[width=\linewidth]{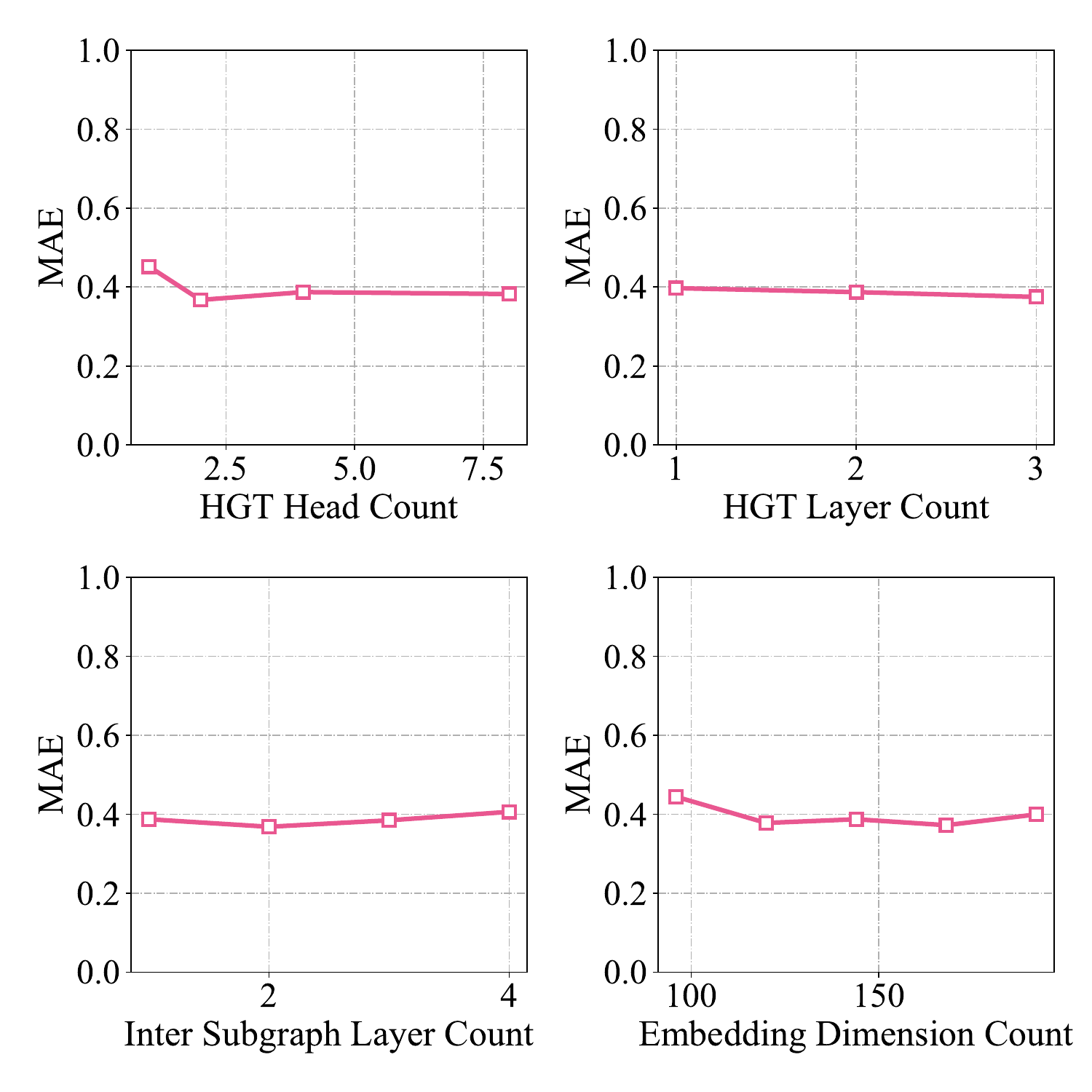} 
        \caption{Chicago.}  
        \label{fig:chi_parameter}  
    \end{minipage}  
    \label{fig:parameters}  
\end{figure}

The key parameters that affect the performance of our model include the number of layers and attention heads in the node encoding model, the number of layers in the inter-subgraph model, and the dimensionality of the region embeddings. As shown in the Figure \ref{fig:nyc_parameter} and Figure \ref{fig:chi_parameter}, the parameter settings in HGT have a minimal impact on the overall framework's performance. We configure the node encoding model with 2 layers and set the number of attention heads to 4. For the inter-subgraph model, we opted for a single layer, as increasing the number of layers led to redundant learning of urban region information, which negatively impacted performance.
  
In terms of the region embedding dimensionality, larger embeddings do not always lead to better performance due to the risk of overfitting and the inherent complexity of the data. Therefore, we chose a dimensionality of 144, which we found to offer a good trade-off between expressiveness and generalization. These settings allow our model to efficiently capture the relevant spatial and relational features of urban regions while maintaining scalability across different datasets.

\subsection{Appendix of implementation details} \label{app:implement}
Each city has its own token graph base, which is determined by the available data for that city. The types of tokens in each city's base vary depending on the data, as detailed in Table~\ref{tab:token_types}. This table presents the number of relevant token types for each city.
\begin{table}[h]
  \centering
  \small
  \resizebox{0.48\textwidth}{!}{ 
      \begin{tabular}{c|c|ccc}
        \toprule
        Type &Data & NYC & CHI & SZ\\
        \midrule
        \multirow{3}[1]{*}{spatial token}&
        POI & 147080& 112989& 228424 \\
        &Road & 110919& 139850& 43985 \\
        &Junction & 62627& 66321& 87932 \\
        \midrule
        \multirow{4}[1]{*}{virtual token}&
        POI Top Category & 334& 166& 17 \\
        &POI Sub Category & 1100& - & 108 \\
        &POI Tags & 1057& - & 770 \\
        &Road Category & 6& 6& 37 \\
        \bottomrule
      \end{tabular}
  }
  \caption{Token types and counts in token graphs.}
  \label{tab:token_types}
\end{table}

Additionally, each city has its own data schema, denoted as $\mathcal{D}$, which outlines the relationships and entities specific to that city. Table~\ref{tab:data_schema} provides an overview of the data in $\mathcal{D}$ of each city, including the size of each data set $D_{t_1 \rightarrow t_2}$.

\begin{table}[h]
  \centering
  \small
  \resizebox{0.48\textwidth}{!}{ 
      \begin{tabular}{c|c|ccc}
        \toprule
        $t_1$ & $t_2$ & NYC & CHI & SZ\\
        \midrule
        poi&poi& $4.0 \times 10^4$ & - & - \\
        poi&poi\_brand& $1.1 \times 10^4$& $1.1 \times 10^4$& - \\
        poi&poi\_topcategory & $1.5 \times 10^5$& $1.1 \times 10^5$& $2.3 \times 10^5$\\
        poi&poi\_subcategory&$1.3 \times 10^5$&-&$2.3 \times 10^5$\\
        poi&poi\_categorytags&$1.9 \times 10^5$&-&$2.3 \times 10^5$\\
        road&junction&$1.1 \times 10^5$& $7.2 \times 10^4$& $4.4 \times 10^4$\\
        road&junction&$1.1 \times 10^5$& $7.2 \times 10^4$ &$4.4 \times 10^4$\\
        road&road\_category &$1.1 \times 10^5$& $1.4 \times 10^5$& $4.4 \times 10^4$ \\
        junction&junction\_category & $6.2 \times 10^4$& $6.6 \times 10^4$& - \\
        junction&road &$2.2 \times 10^5$& - & -\\
        \bottomrule
      \end{tabular}
  }
  \caption{Relation types and counts in token graphs.}
  \label{tab:data_schema}
\end{table}

\subsection{Performance of elastic region representation in Shenzhen} \label{app:shenzhen}
We also conduct experiments on Shenzhen, as shown in Table~\ref{tab:shenzhen}. In this case, the tasks include population and nighttime light prediction. The results demonstrate that our elastic urban region representation (EURR) achieves state-of-the-art (SOTA) performance, further validating the effectiveness of the proposed framework in real-world scenarios.
\begin{table}[h]
  \vspace{0.3cm}
  \resizebox{0.5\textwidth}{!}{%
  \begin{tabular}{c|ccc|ccc}
    \toprule
    \multirow{2}[1]{*}{Model} & \multicolumn{3}{c|}{Nightlight prediction~(SZ)} & \multicolumn{3}{c}{Population Prediction~(SZ)}\\
    &
    \multicolumn{1}{c}{MAE} & \multicolumn{1}{c}{RMSE} & \multicolumn{1}{c|}{$R^2$} &
    \multicolumn{1}{c}{MAE} & \multicolumn{1}{c}{RSME} & \multicolumn{1}{c}{$R^2$} \\
    \toprule
    TransR & 0.794& 0.973& 0.052& 0.739& 0.870& 0.243\\
    node2vec & 0.864& 0.994& 0.013& 0.967& 1.220& -0.489\\
    GAE & 0.848& 1.003& -0.006& 0.888& 0.995& 0.010\\  
    \toprule
    HREP& 0.796& 0.952& 0.093& 0.623& 0.835& 0.303\\  
    MGFN& 0.855& 1.014& -0.028& 0.694& 0.895& 0.198\\  
    MVURE& 0.822& 0.976& 0.047& 0.629& 0.830& 0.310\\  
    ReCP & 0.820& 0.993& 0.013& \underline{0.604}& 0.808& 0.347\\
    \toprule
    BPURF-mini & \underline{0.588}& \underline{0.730}& \underline{0.466}& 0.613& \underline{0.769}& \textbf{0.408}\\
    BPURF & \textbf{0.421}&  \textbf{0.526}& \textbf{0.722}&  \textbf{0.547}&  \textbf{0.769}&  \underline{0.407}\\
    \bottomrule
  \end{tabular}
  }
  \caption{EURR results in ShenZhen.}
  \label{tab:shenzhen}
  \vspace{-0.1cm}
\end{table}

\subsection{Appendix of performance on static region embedding.} \label{app:static}
We evaluate our model on traditional region embedding tasks, where region boundaries are fixed during both training and inference. For the baseline models, we use their predefined region partitions for training and validation. Our model is trained on dynamic regions, and we use boundary prompts to perform inference on these fixed regions.

\begin{table*}[htbp]
  \small
  \vspace{-0.3cm}

  \resizebox{1\textwidth}{!}{%
  \begin{tabular}{c|ccc|ccc|ccc|ccc}
    \toprule
    \multirow{3}[1]{*}{Model} & \multicolumn{6}{c|}{New York City~(NYC)} & \multicolumn{6}{c}{Chicago~(CHI)} \\
    \cmidrule{2-13}
     & \multicolumn{3}{c|}{Crime Prediction} & \multicolumn{3}{c|}{Check-in Prediction} & \multicolumn{3}{c|}{Crime Prediction} & \multicolumn{3}{c}{Crash Prediction}\\
    \cmidrule{2-13}
    &
    \multicolumn{1}{c}{MAE} & \multicolumn{1}{c}{RMSE} & \multicolumn{1}{c|}{$R^2$} &
    \multicolumn{1}{c}{MAE} & \multicolumn{1}{c}{RMSE} & \multicolumn{1}{c|}{$R^2$} &
    \multicolumn{1}{c}{MAE} & \multicolumn{1}{c}{RMSE} & \multicolumn{1}{c|}{$R^2$} &
    \multicolumn{1}{c}{MAE} & \multicolumn{1}{c}{RMSE} & \multicolumn{1}{c}{$R^2$}\\
    \toprule
    TransR & 0.753 & 1.007 & -0.014 & 0.654 & 0.984 & 0.031 & 0.692 & 1.018 & -0.037 & 0.604 & 1.000 & 0.001  \\
    node2vec & 0.795 & 1.030 & -0.062 & 0.590 & 0.841 & 0.292 & 0.845 & 1.184 & -0.402 & 0.754 & 1.102 & -0.214 \\
    GAE & 0.758 & 1.015 & -0.030 & 0.678 & 1.009 & -0.018 & 0.701 & 1.028 & -0.057 & 0.604 & 1.011 & -0.022 \\
    \midrule
    HREP & 0.536 & 0.705 & 0.503 & 0.472 & 0.683 & 0.534 & 0.698 & 1.015 & -0.030 & 0.569 & 0.902 & 0.186  \\
    MGFN & 0.605 & 0.778 & 0.395 & 0.483 & 0.652 & 0.575 & 0.731 & 1.059 & -0.122 & 0.638 & 1.054 & -0.110 \\
    MVURE & 0.487 &\underline{0.661} & \underline{0.563} & 0.421 & 0.628 & 0.606 & 0.691 & 0.973 & 0.054 & 0.557 & 0.858 & 0.263 \\
    ReCP & 0.512 & 0.696 & 0.516 & 0.373 & 0.641 & 0.589 & 0.687 & 0.990 & 0.020 & 0.529 & 0.892 & 0.205 \\
    GURPP & \underline{0.498} & 0.705 &0.503 & \underline{0.317} &\underline{0.509} &\underline{ 0.741} &\underline{0.627} & \underline{0.924} & \underline{0.159} & \underline{0.473} &\underline{0.979} & \underline{0.032} \\
    \midrule 
    BPURF & \textbf{0.468} & \textbf{0.624} & \textbf{0.611} & \textbf{0.316} & \textbf{0.507} & \textbf{0.743} & \textbf{0.548} & \textbf{0.778} & \textbf{0.404} & \textbf{0.456} & \textbf{0.733} & \textbf{0.456} \\
    \bottomrule
\end{tabular}
}
\caption{Main performance comparison of different models on static regions across NYC and CHI datasets. The best results for each metric are highlighted in bold, the second-best are underlined.}
  \label{tab:main_performance_result_static}
\end{table*}

\subsection{Appendix of extra ablation study} \label{app:ablation}
As shown in Table \ref{tab:main_performance_result_static}, even when evaluated on fixed regions that were not encountered during training, our model achieves performance that is better than existing state-of-the-art (SOTA) methods. This highlights the strong generalization capability of our framework, which was originally designed for dynamic regions, across various urban region scenarios.

\begin{figure}[h]
    \centering
    \includegraphics[width=\linewidth]{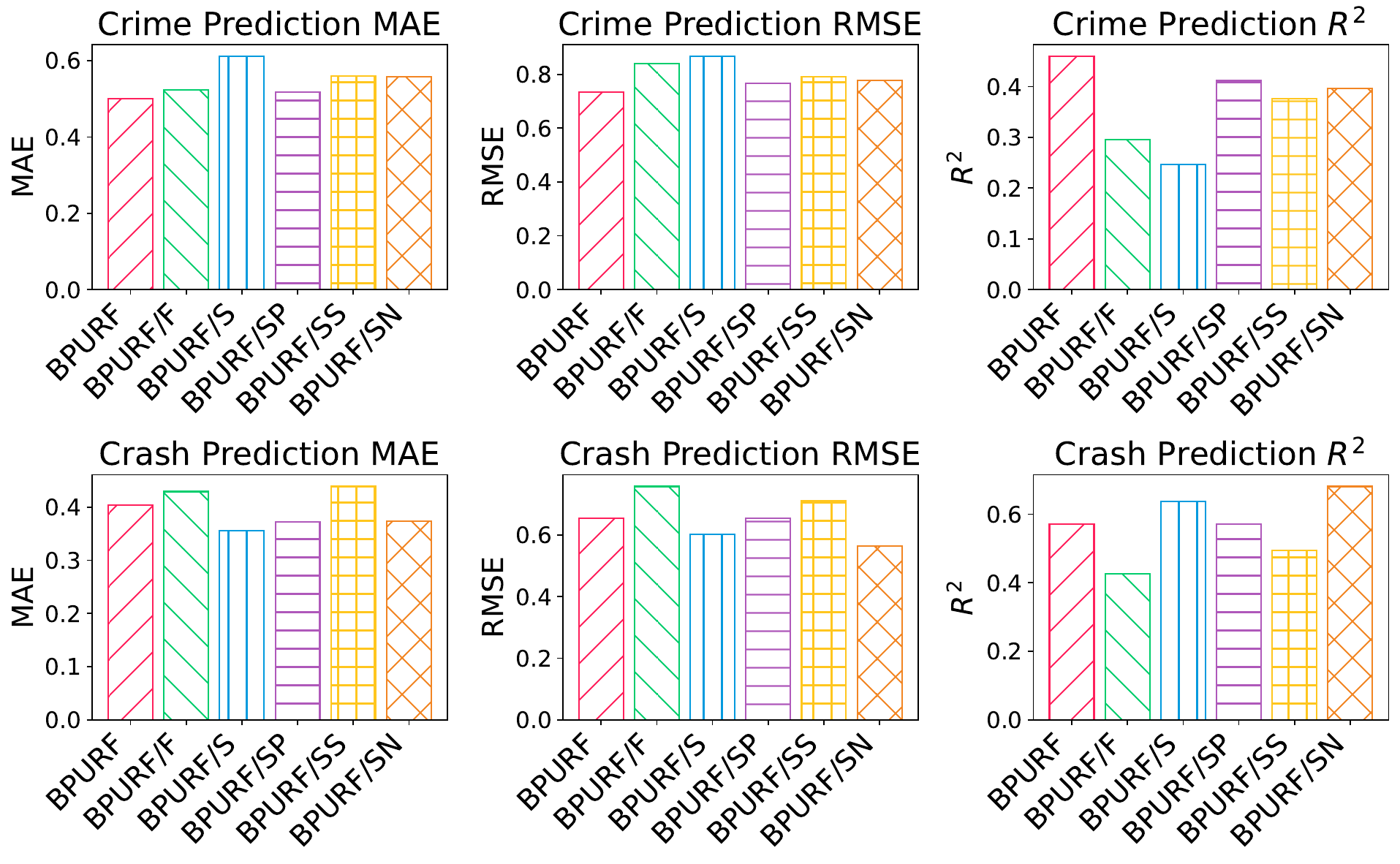}
    \caption{Results of ablation studies in Chicago.}
    \label{fig:Ablation_chi}
    \vspace{-15pt}
\end{figure}

\subsubsection{Ablation study in Chicago}
As shown in Figure \ref{fig:Ablation_chi}, we also conducted an ablation study on the Chicago dataset. The results indicate that removing certain modules leads to a noticeable decline in performance.

\end{document}